\documentclass[runningheads]{llncs}

\usepackage[T1]{fontenc}
\usepackage{graphicx}
\usepackage{float}

\usepackage[utf8]{inputenc}
\usepackage{indentfirst}
\usepackage{hyperref}
\usepackage{booktabs}

\usepackage{subfigure}

\begin{document}
\title{Support Vector Machine-Based Burnout Risk Prediction with an Interactive Interface for Organizational Use}

\titlerunning{Burnout Risk Prediction with SVM and Interactive Interface}

\author{Bruno W. G. Teodosio\inst{1} \and Mário J. O. T. Lira\inst{1} \and Pedro H. M. Araújo\inst{1,4} \and Lucas R. C. Farias\inst{1,2,3}}

\authorrunning{Teodosio et al.}

\institute{
Universidade Católica de Pernambuco (UNICAP), Recife, Pernambuco, Brazil \\
\email{lucas.farias@unicap.br, brunowashingtong@gmail.com, mario.846606@unicap.br, pedro.araujo@unicap.br}
\and
Centro de Informática – Universidade Federal de Pernambuco (UFPE), Recife, Pernambuco, Brazil
\and
Centro de Estudos e Sistemas Avançados do Recife (CESAR School), Recife, Pernambuco, Brazil
\and
Universidade de Pernambuco (UPE), Recife, Pernambuco, Brazil
}

\maketitle              % typeset the header of the contribution
\begin{abstract}
Burnout is a psychological syndrome marked by emotional exhaustion, depersonalization, and reduced personal accomplishment, with a significant impact on individual well-being and organizational performance. This study proposes a machine learning approach to predict burnout risk using the HackerEarth Employee Burnout Challenge dataset. Three supervised algorithms were evaluated: nearest neighbors (KNN), random forest, and support vector machine (SVM), with model performance evaluated through 30-fold cross-validation using the determination coefficient (R²). Among the models tested, SVM achieved the highest predictive performance (R² = 0.84) and was statistically superior to KNN and Random Forest based on paired $t$-tests. To ensure practical applicability, an interactive interface was developed using Streamlit, allowing non-technical users to input data and receive burnout risk predictions. The results highlight the potential of machine learning to support early detection of burnout and promote data-driven mental health strategies in organizational settings.


\end{abstract}

\section{Introduction}

Burnout is a psychological syndrome defined by emotional exhaustion, depersonalization, and a diminished sense of personal achievement. It is particularly prevalent in high-stress professional environments and has significant implications for both individual well-being and organizational effectiveness~\cite{pillai2024using,kaggle-burnout}. Early identification of individuals at risk is critical to supporting mental health and sustaining workforce productivity.

In response to this challenge, we investigate the application of supervised machine learning to predict burnout risk using the HackerEarth Employee Burnout Challenge dataset~\cite{kaggle-burnout}. This dataset includes diverse employee profiles, workplace conditions, and mental health indicators, providing a valuable foundation for developing predictive models. Although the data is not tailored to a specific domain, it captures features relevant across organizational contexts.

This study evaluates the performance of three widely used regression algorithms such as K-Nearest Neighbors (KNN), Random Forest, and Support Vector Machine (SVM) using the coefficient of determination ($R^2$) and the cross-validation of 30 times \cite{lara2021}. Statistical tests were applied to assess the significance of observed differences. Among the evaluated models, SVM achieved the highest predictive accuracy and was statistically superior to the alternatives. To ensure the practical applicability of the model, an interactive interface was developed using Streamlit, allowing non-technical users to make predictions and interpret results with ease.

An interactive interface was developed using Streamlit to facilitate real-world adoption, allowing non-technical users to generate burnout risk estimates based on input parameters. This integration of prediction and accessibility highlights the potential of machine learning to support early detection of mental health issues and promote proactive, data-driven interventions in organizational settings.

The remainder of this paper is organized as follows: Section 2 reviews relevant literature on burnout prediction and machine learning applications in mental health. Section 3 presents the dataset, preprocessing steps, learning models, and evaluation metrics. Section 4 discusses the experimental results and statistical validation. Section 5 describes the deployment of the interactive interface. Finally, Section 6 summarizes the contributions and suggests directions for future work.

\section{Related Work}
\label{sec:related-work}
Burnout syndrome, first conceptualized by Freudenberger~\cite{freudenberger1974}, is characterized by emotional exhaustion, depersonalization, and a diminished sense of personal accomplishment. It is associated with adverse outcomes for both individual well-being and organizational performance, and has become an increasingly important subject in occupational health research.

Pustulka-Piwnik et al.~\cite{pustulka2014} examined the prevalence of burnout among physiotherapists, identifying demographic and organizational risk factors, such as workload and lack of autonomy. Their findings reinforced the importance of identifying high-risk profiles to guide preventive interventions in workplace environments.

With the growing availability of behavioral and occupational datasets, machine learning (ML) has gained prominence as a tool for mental health analysis. Adapa et al.~\cite{adapa2022} applied supervised learning algorithms to predict burnout in healthcare professionals, demonstrating that data-driven models can uncover complex relationships between individual characteristics and burnout symptoms.

Lara-González et al.~\cite{lara2021} conducted a comparative evaluation of ML algorithms, including KNN, SVM, Random Forest, and neural networks, for the prediction of burnout. Their results highlighted the superior performance of SVM and Random Forest in identifying latent risk profiles, particularly due to their capacity to handle nonlinear interactions and heterogeneous feature spaces.

Tang et al.~\cite{tang2023} further contributed to the field by employing counterfactual explanations to improve the interpretability of the model. Their approach provided actionable insights into the variables most associated with burnout among shift workers, enabling more transparent and explainable predictive systems.

These studies collectively illustrate the growing relevance of ML methods in mental health research, especially for predictive and preventive applications in occupational settings. Building upon this foundation, the present study evaluates the performance of KNN, SVM, and Random Forest models for predicting burnout risk using real-world employee data. The objective is not only to identify the most accurate model but also to provide a practical tool for supporting early intervention strategies in corporate environments.

\section{Methodology}
\label{sec:methodology}

This Section describes the methodological steps followed in the development, evaluation, and validation of the burnout prediction model.

\subsection{Dataset and Features}

We used the public dataset \textit{"Are Your Employees Burning Out?"}, available on Kaggle~\cite{kaggle-burnout}, containing 22,750 records with employee information relevant to burnout analysis~\cite{dataset-analysis-approach,survival-analysis-mental-fatigue,eeg-mental-fatigue-detection}. The dataset includes features grouped into four categories:

\begin{itemize}
    \item \textbf{Identification and Tenure:} Employee ID and date of joining.
    \item \textbf{Demographics and Work Setup:} Gender, company type (Product/Service), and availability of remote work (WFH).
    \item \textbf{Work Characteristics:} Designation and resource allocation (scale from 1 to 10).
    \item \textbf{Mental Health Indicators:} Mental fatigue score (0–10) and burn rate (0–1), the latter used as the target variable.
\end{itemize}

\subsection{Data Preprocessing}

To prepare the dataset for modeling, the following steps were applied:

\begin{itemize}
    \item \textbf{Missing Values:} Median imputation was applied to numerical fields with missing values (\textit{Resource Allocation}, \textit{Mental Fatigue Score}, and \textit{Burn Rate}).
    \item \textbf{Encoding:} Categorical variables (\textit{Gender}, \textit{Company Type}, and \textit{WFH Setup}) were one-hot encoded.
    \item \textbf{Normalization:} Features were standardized using \textit{StandardScaler} (zero mean, unit variance), which is required for distance- and margin-based algorithms such as KNN and SVM.
\end{itemize}

\subsection{Models and Evaluation Metrics}

Three supervised regression algorithms were evaluated: K-Nearest Neighbors (KNN), Random Forest, and Support Vector Machine (SVM). These were selected based on prior evidence of effectiveness in burnout prediction~\cite{lara2021}.

The primary performance metric was the coefficient of determination ($R^2$), which quantifies how well the model explains variance in the continuous target variable. A value of $R^2=1$ indicates a perfect fit; values near zero or negative reflect poor explanatory power.

\subsubsection{K-Nearest Neighbors (KNN)}

KNN predicts values based on proximity to $k$ training instances. While simple and non-parametric, its performance depends on the selection of $k$ and is sensitive to noise and outliers. KNN yielded $R^2 = 0.8371$, capturing local patterns in variables like \textit{Mental Fatigue Score} and \textit{Resource Allocation}.

\subsubsection{Random Forest}

Random Forest combines multiple decision trees trained on bootstrap samples with random feature selection at each split. It achieved $R^2 = 0.8363$, identifying key predictors while offering model interpretability via feature importance. Despite strong performance, it was slightly outperformed by SVM and is more computationally intensive.

\subsubsection{Support Vector Machine (SVM)}

SVM with a radial basis function (RBF) kernel was used to capture nonlinear relationships in the data. It achieved the best performance ($R^2 = 0.8479$), showing the capacity to mitigate overfitting and the ability to model subtle interactions between features. The key hyperparameters ($C$, $\epsilon$, and $\gamma$) were tuned to balance generalization and precision.

\subsection{Cross-Validation Protocol}

We used 30-fold cross-validation for model evaluation and to reduce sensitivity to random data splits. In each iteration, the model was trained on 29 folds and validated on the remaining one. The final $R^2$ was calculated as the mean across folds. This strategy ensures full data usage for both training and validation, improving statistical reliability of the comparisons.

\subsection{Tools and Libraries}

The entire pipeline was developed using open-source Python libraries, including \textit{Scikit-Learn}, \textit{Pandas}, \textit{NumPy}, \textit{Seaborn}, \textit{Matplotlib}, \textit{Joblib} (for model persistence), and \textit{Streamlit} (for deployment of the predictive interface).

\section{Results and Discussion}
\label{sec:results-discussion}

\subsection{Exploratory Analysis and Data Quality}

The dataset consists of 22,750 records and nine features encompassing personal, professional, and mental health attributes. Exploratory Data Analysis (EDA) was conducted to assess variable distributions, detect missing values, and identify potential correlations.

\begin{figure}[htbp]
    \centering
    \includegraphics[width=0.8\linewidth]{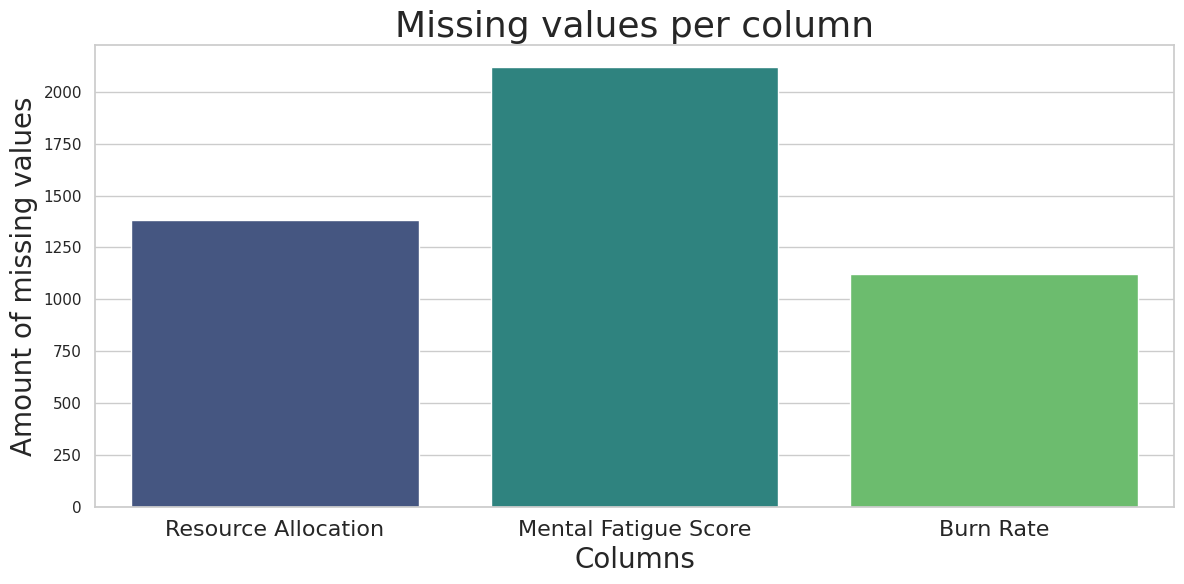}
    \caption{Bar chart showing the number of missing values per feature in the dataset. Only \textit{Resource Allocation}, \textit{Mental Fatigue Score}, and \textit{Burn Rate} presented missing data; all other variables were complete.}
    \label{fig:missing}
\end{figure}

Three variables contained missing values—\textit{Resource Allocation}, \textit{Mental Fatigue Score}, and \textit{Burn Rate}. These records were excluded to preserve data consistency (Figure~\ref{fig:missing}).

\subsection{Feature Distributions and Correlation Patterns}

\begin{figure}[htbp]
    \centering
    \includegraphics[width=0.7\linewidth]{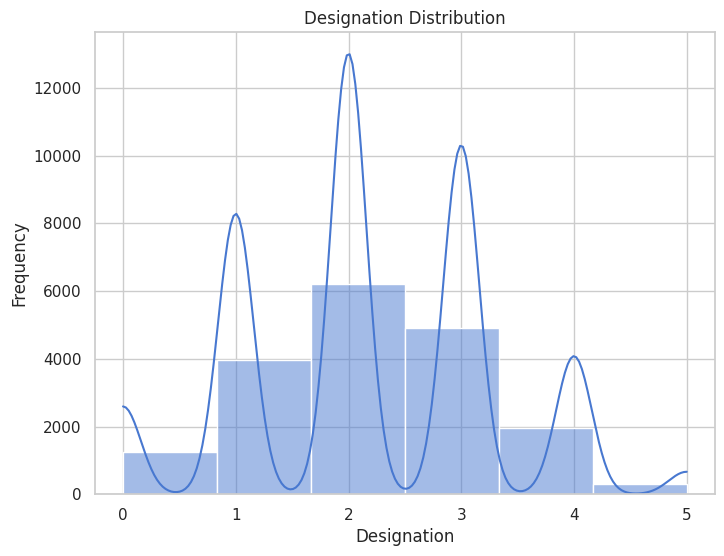}
    \caption{Distribution of the \textit{Designation} variable, illustrating the frequency of employee positions across hierarchical levels. The discrete pattern reflects structured job roles within the organization.}
    \label{fig:designacao}
\end{figure}

\begin{figure}[htbp]
    \centering
    \includegraphics[width=0.7\linewidth]{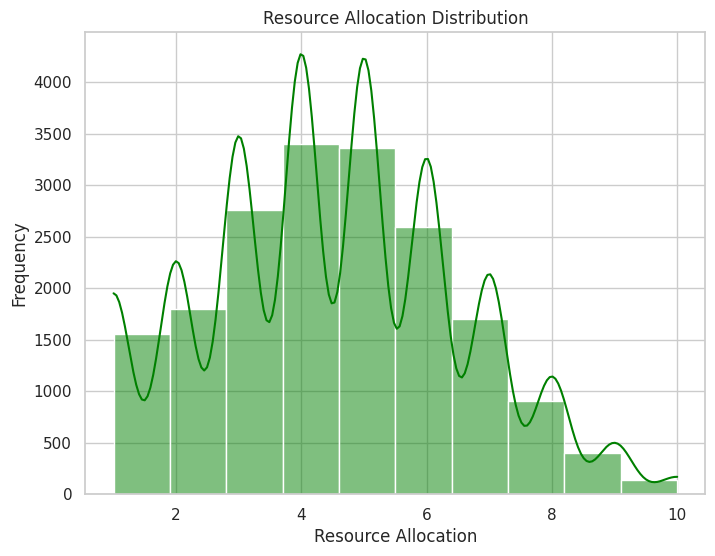}
    \caption{Distribution of the \textit{Resource Allocation} variable across the dataset. Most values are concentrated between 4 and 6, suggesting a moderate and consistent allocation of resources among employees.}
    \label{fig:recursos}
\end{figure}

The variable \textit{Designation} exhibited a discrete distribution aligned with job hierarchy (Figure~\ref{fig:designacao}), while \textit{Resource Allocation} showed a uniform concentration between 4 and 6 (Figure~\ref{fig:recursos}). Both \textit{Mental Fatigue Score} and \textit{Burn Rate} presented symmetrical distributions and were highly correlated (Pearson $r = 0.94$).

\begin{figure}[htbp]
    \centering
    \includegraphics[width=0.7\linewidth]{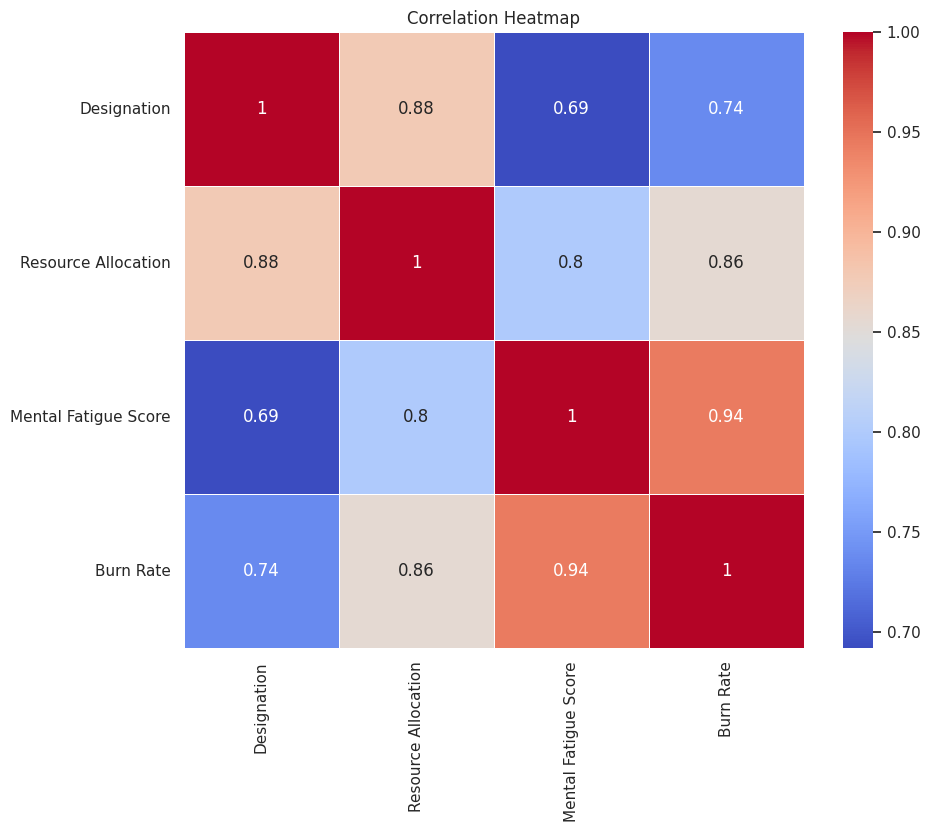}
    \caption{Correlation heatmap of key numerical features in the dataset. The highest correlation was observed between \textit{Mental Fatigue Score} and \textit{Burn Rate} ($r = 0.94$), suggesting a strong association between psychological strain and burnout risk.}
    \label{fig:correlacao}
\end{figure}

\begin{figure}[htbp]
    \centering
    \includegraphics[width=0.8\linewidth]{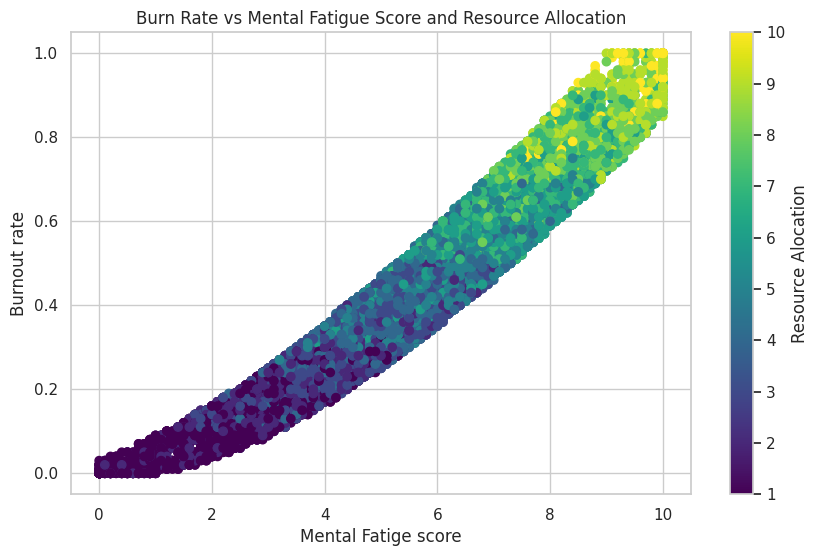}
    \caption{Multivariate plot illustrating the relationship between \textit{Burn Rate}, \textit{Mental Fatigue Score}, and \textit{Resource Allocation}. Higher levels of mental fatigue and resource allocation are associated with increased burnout rates, as shown by the upward trend and color gradient.}
    \label{fig:dispersao}
\end{figure}

\subsection{Group-Level Insights}

\begin{figure}[htbp]
    \centering
    \includegraphics[width=0.8\linewidth]{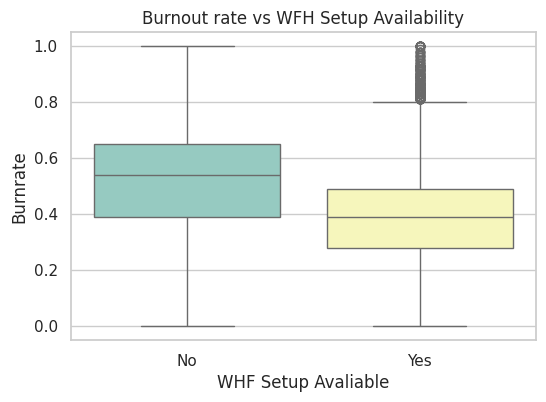}
    \caption{Boxplot of burnout rates segmented by remote work availability. Employees without access to a work-from-home (WFH) setup exhibited higher median burnout levels, suggesting a potential protective effect of remote work in reducing psychological strain.}
    \label{fig:home_office}
\end{figure}

Categorical comparisons revealed that employees without remote work access presented higher burnout medians (0.6 vs. 0.4) (Figure~\ref{fig:home_office}). Gender-based differences were marginal, and company type (Product vs. Service) did not significantly impact burnout distribution.

\subsection{Statistical Validation and PCA}

\begin{figure}[htbp]
    \centering
    \includegraphics[width=0.8\linewidth]{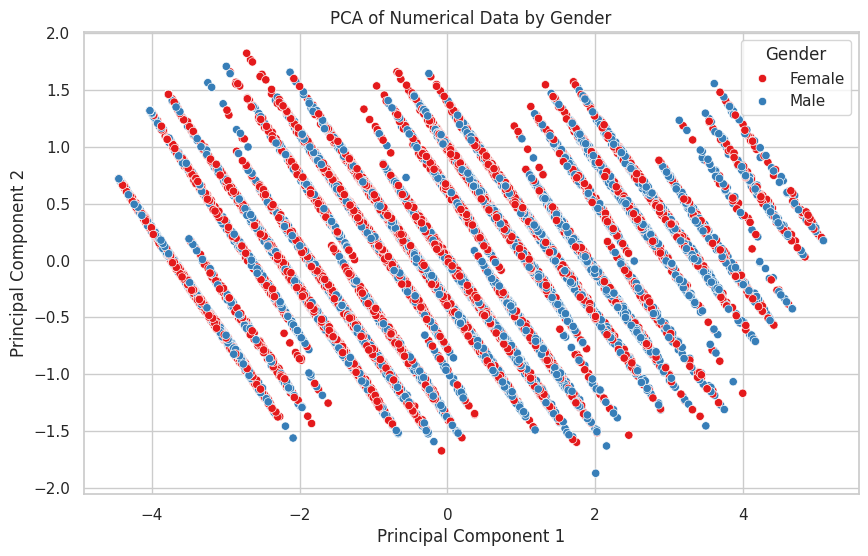}
    \caption{Principal Component Analysis (PCA) projection of numerical features, colored by gender. The overlap between male and female data points indicates no evident clustering or separation, suggesting that gender does not strongly differentiate the observed burnout-related patterns in the dataset.}
    \label{fig:pca}
\end{figure}

Normality tests confirmed non-Gaussian distributions. T-tests indicated a statistically significant difference in burnout rates between remote and on-site workers ($p < 0.01$). Principal Component Analysis (PCA) did not reveal clear group separation by gender (Figure~\ref{fig:pca}).

\subsection{Model Performance Comparison}

Three supervised models—KNN, Random Forest, and SVM—were trained and evaluated using 30-fold cross-validation. Table~\ref{tab:r2_summary} summarizes the mean $R^2$ scores.

\begin{table}[htbp]
\caption{Average $R^2$ scores obtained from 30-fold cross-validation for each evaluated model. SVM achieved the highest mean performance, followed closely by KNN and Random Forest.}
\begin{center}
\begin{tabular}{c|c}
\hline
\textbf{Model} & \textbf{Average $R^2$} \\
\hline
K-Nearest Neighbors & 0.83 \\
Support Vector Machine & \textbf{0.84} \\
Random Forest & 0.83 \\
\hline
\end{tabular}
\label{tab:r2_summary}
\end{center}
\end{table}

Paired t-tests (Table~\ref{tab:stat_tests}) revealed that SVM significantly outperformed KNN ($p = 0.036$) and Random Forest ($p = 0.027$). No significant difference was found between KNN and Random Forest ($p = 0.869$).

\begin{table}[htbp]
\caption{Pairwise $t$-tests comparing model performance based on $R^2$ scores. Statistically significant differences ($\alpha = 0.05$) are highlighted in bold. SVM outperformed both KNN and Random Forest with $p < 0.05$, while no significant difference was observed between KNN and Random Forest.}
\begin{center}
\begin{tabular}{c|c|c}
\hline
\textbf{Comparison} & \textbf{t-stat} & \textbf{p-value} \\
\hline
KNN vs SVM & -2.49 & \textbf{0.036} \\
KNN vs Random Forest & 0.16 & 0.869 \\
SVM vs Random Forest & 2.69 & \textbf{0.027} \\
\hline
\end{tabular}
\label{tab:stat_tests}
\end{center}
\end{table}

\subsection{Model Saving and Deployment}

After selecting SVM as the final model, it was serialized using \texttt{Joblib} for deployment. The model and \textit{StandardScaler} were stored in separate \texttt{.pkl} files to preserve preprocessing consistency.

\begin{itemize}
    \item \textbf{Training and Parameters:} The SVM was trained with an RBF kernel, using $C = 1.0$ and $\epsilon = 0.1$ for regularization and error tolerance.
    \item \textbf{Integration:} The stored files were integrated into a Streamlit-based interface, enabling practical application by non-technical users.
\end{itemize}

\section{Interactive Interface Using Streamlit}
\label{sec:implementation}

To facilitate real-world use of the predictive model, an interactive web interface was developed using the \textit{Streamlit} library. The interface allows users, such as managers and HR professionals, to input relevant employee data and instantly receive an estimated risk of burnout. This section details the design and functionality of the deployed interface.

\subsection{Purpose and Accessibility}

The primary goal of the interface is to democratize access to the predictive model by offering a simple and intuitive user experience. Designed for non-technical users, the system enables real-time monitoring of employee mental health without requiring knowledge of machine learning or data preprocessing. The interface bridges the gap between technical modeling and practical decision-making in organizational settings.

\subsection{User Interaction and Components}

The interface consists of interactive elements that collect input from users and provide dynamic feedback:

\begin{itemize}
    \item \textbf{Sliders for Numerical Variables:} Users can set values for continuous features such as \textit{Mental Fatigue Score}, \textit{Resource Allocation}, and \textit{Designation} using intuitive slider widgets.

    \item \textbf{Dropdown Selectors for Categorical Variables:} Features such as \textit{Gender}, \textit{Company Type}, and \textit{WFH Setup Available} are selected from predefined options via dropdown menus.

    \item \textbf{Automatic Normalization:} Before prediction, all inputs are automatically standardized using the same \textit{StandardScaler} applied during model training. This ensures consistency and maintains model accuracy.
\end{itemize}

\subsection{Prediction Pipeline}

Once the input data is provided, the prediction process follows three core steps:

\begin{itemize}
    \item \textbf{Data Preparation:} Inputs are collected and preprocessed, including normalization using the previously saved \textit{StandardScaler} object.

    \item \textbf{Model Inference:} The trained SVM model, stored in \texttt{modelo\_svm\_vencedor.pkl}, is loaded and used to compute the predicted burnout rate.

    \item \textbf{Result Display:} The prediction, ranging from 0 to 1, is displayed immediately on the interface, providing an interpretable risk score.
\end{itemize}

\subsection{Deployment and Access}

\begin{figure}[htbp]
    \centering
    \includegraphics[width=0.3\linewidth]{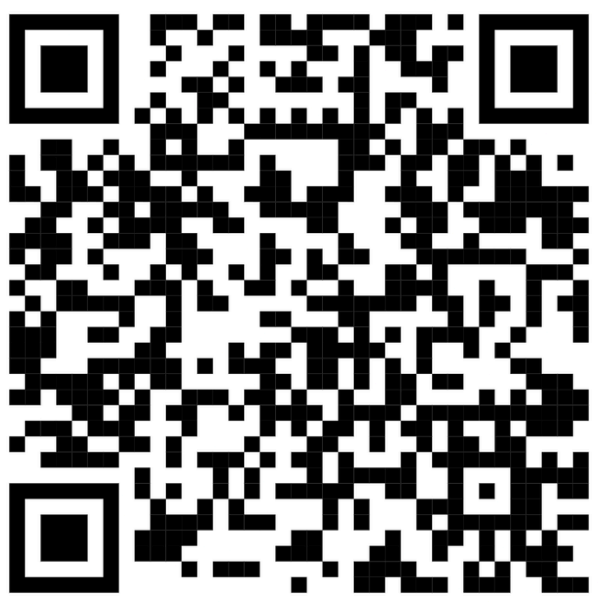} 
    \caption{QR code linking to the online deployment of the Streamlit-based burnout prediction application. The interface allows users to input employee data and receive real-time burnout risk estimates via a publicly accessible platform.}
    \label{fig:qr_code}
\end{figure}

The full application was deployed online via \textit{Streamlit Cloud} and is accessible at:  
\url{https://employee-burnout-svm.streamlit.app/}. A QR code linking to the platform is provided in Figure~\ref{fig:qr_code} for ease of access.

\subsection{Practical Utility}

The deployed system supports use in various professional contexts by offering a no-code, real-time prediction tool. It enables organizations to monitor burnout risk in a privacy-preserving and non-invasive manner, supporting preventive strategies and evidence-based decision-making. The interface can also be easily adapted to different data profiles, broadening its applicability across industries.

\section{Conclusion}
\label{sec:conclusion}

This study developed a predictive model to assess the risk of employee burnout using supervised learning techniques applied to the “Are Your Employees Burning Out?” dataset from Kaggle. Among the evaluated models, the Support Vector Machine (SVM) with a radial kernel and optimized parameters achieved the best performance, demonstrating a strong ability to identify complex patterns associated with burnout risk. As a practical contribution, an interactive interface was built using Streamlit, allowing users without technical expertise to input data and receive intuitive predictions, facilitating the use of the model in corporate settings for early detection and prevention.

Despite its effectiveness, the SVM model presented some limitations, particularly its sensitivity to data normalization and hyperparameter tuning. Since SVM relies on distance-based calculations, inconsistent feature scaling can compromise prediction accuracy. Additionally, the model's performance depends heavily on the calibration of parameters such as C and epsilon, which influence regularization and error tolerance. Future work should explore more refined tuning methods and consider incorporating additional features—such as employee performance or job satisfaction metrics—to enhance the model’s predictive power and practical applicability.

Future research should explore alternative algorithms, such as Neural Networks or explainability-focused models, to deepen the understanding of factors contributing to burnout. Incorporating additional data sources may enhance the model’s generalization and predictive accuracy, while optimization techniques like Grid Search or Random Search can improve hyperparameter tuning. This study reinforces the potential of Machine Learning as a valuable and accessible tool for monitoring mental health in the workplace. The proposed strategies and methods are adaptable to other contexts, supporting a data-driven approach to promoting employee well-being.

\bibliography{bibliography}

\end{document}